\documentclass[conference,a4paper]{IEEEtran}
\IEEEoverridecommandlockouts
\usepackage{cite}
\usepackage{amsmath,amssymb,amsfonts}
\usepackage{algorithmic}
\usepackage{graphicx}
\usepackage{textcomp}
\usepackage{xcolor}
\usepackage{subcaption}
\usepackage{caption}
\DeclareCaptionJustification{justified}{\justifying}
\usepackage[raggedrightboxes]{ragged2e}
\usepackage{algorithm}
\usepackage{algorithmic}
\usepackage{float}
\usepackage{multirow}

\def\BibTeX{{\rm B\kern-.05em{\sc i\kern-.025em b}\kern-.08em
    T\kern-.1667em\lower.7ex\hbox{E}\kern-.125emX}}
\begin{document}

\title{Fast-OMRA: Fast Online Motion Resolution Adaptation for Neural B-Frame Coding\\
}

\author{\IEEEauthorblockN{Sang NguyenQuang\textsuperscript{1}, Zong-Lin Gao\textsuperscript{1}, Kuan-Wei Ho\textsuperscript{1}, Xiem HoangVan\textsuperscript{2}, Wen-Hsiao Peng\textsuperscript{1}}
\IEEEauthorblockA{{\textsuperscript{1}Department of Computer Science, National Yang Ming Chiao Tung University, Hsinchu, Taiwan} \\
\textsuperscript{2}Faculty of Electronics and Telecommunications, VNU University of Engineering and Technology, Hanoi, Vietnam \\}
}

\maketitle
\begin{abstract}
Most learned B-frame codecs with hierarchical temporal prediction suffer from the domain shift issue caused by the discrepancy in the Group-of-Pictures (GOP) size used for training and test. As such, the motion estimation network may fail to predict large motion properly. One effective strategy to mitigate this domain shift issue is to downsample video frames for motion estimation. However, finding the optimal downsampling factor involves a time-consuming rate-distortion optimization process. This work introduces lightweight classifiers to determine the downsampling factor. To strike a good rate-distortion-complexity trade-off, our classifiers observe simple state signals, including only the coding and reference frames, to predict the best downsampling factor. We present two variants that adopt binary and multi-class classifiers, respectively. The binary classifier adopts the Focal Loss for training, classifying between motion estimation at high and low resolutions. Our multi-class classifier is trained with novel soft labels incorporating the knowledge of the rate-distortion costs of different downsampling factors. Both variants operate as add-on modules without the need to re-train the B-frame codec. Experimental results confirm that they achieve comparable coding performance to the brute-force search methods while greatly reducing computational complexity.
\end{abstract}



\section{Introduction}
\label{sec:Introduction}


Learned video compression~\cite{dvclu, fvc, mlvc, elfvc} has achieved promising coding performance, with some methods~\cite{dcvcdc, dcvcfm} surpassing traditional codecs like H.266/VVC~\cite{vtm}. However, most research focuses on P-frame coding, leaving B-frame coding~\cite{Wu_2018, murat_lhbdc, iscas_gc24} largely unexplored.

B-frame coding, which leverages the future and past reference frames for temporal prediction, generally outperforms P-frame coding, which relies on uni-directional temporal prediction. However, most state-of-the-art learned B-frame codecs~\cite{bcanf, maskcrt} perform worse than learned P-frame codecs~\cite{hemli, dcvcdc}. 
One root cause is the domain shift issue~\cite{bcanf}. This arises from training learned B-frame codecs on short video sequences with small group-of-pictures (GOP) but applying them to coding large GOPs during inference. For instance, most learned codecs are trained on Vimeo-90K~\cite{vimeo}, where training sequences are only 7 frames long. However, much larger GOPs with 32 frames or more are used for actual encoding. The motion estimation network optimized for small GOPs struggles to estimate well large motion inherent in large GOPs. This results in poor-quality warped frames and thus poor coding performance.

To mitigate this undesirable inconsistency between training and test, several works~\cite{OMRA, MAIF, longvideocoding} have been proposed. Some~\cite{longvideocoding} simply curate their own in-house datasets with long training sequences. However, training B-frame codecs on long sequences is tricky and time consuming. Gao \emph{et al.}~\cite{OMRA} thus propose an Online Motion Resolution Adaptation (OMRA) method, without re-training the codec. It adaptively downsamples the coding and reference frames for motion estimation, turning large motion into small motion when necessary. The resulting optical flow map is then coded and superresolved for warping and prediction. Despite its simplicity, OMRA achieves considerable gains in coding performance. In a similar vein, Yilmaz \emph{et al.}~\cite{MAIF} adjust the motion predictor during inference by downsampling reference frames with four different scale factors and selecting the optimal one based on the quality of the warped frame. In~\cite{OMRA} and~\cite{MAIF}, the optimal downsampling factor has to be searched exhaustively via rate-distortion optimization.   

Built on OMRA, this paper introduces learned classifiers that observe the coding and reference frames to decide the downsampling factor. We first conduct a complexity analysis to understand the complexity characteristics of performing motion estimation at different frame resolutions. We then propose Bi-Class Fast-OMRA and Mu-Class Fast-OMRA. Bi-Class Fast-OMRA is a binary classifier that decides between motion estimation at high and low resolutions. Meanwhile, Mu-Class Fast-OMRA is introduced to skip entirely the search of the best downsampling factor. Both variants operate as add-on modules without the need to re-train the B-frame codec. Experimental results confirm that they achieve comparable coding performance to the brute-force search methods while greatly reducing computational complexity.


The remainder of this paper is organized as follows. Section II describes the proposed framework as well as the construction of our classifiers. Section III assesses the rate-distortion-complexity performance of the proposed methods against several baselines that adopt brute-force search. Section IV concludes this work with major findings.

\section{Proposed Method: Fast-OMRA}
\label{sec:proposed_method}
\begin{figure*}[t]
    \centering
    \vspace{-0.5em}
    \centerline{\includegraphics[width=1\textwidth]{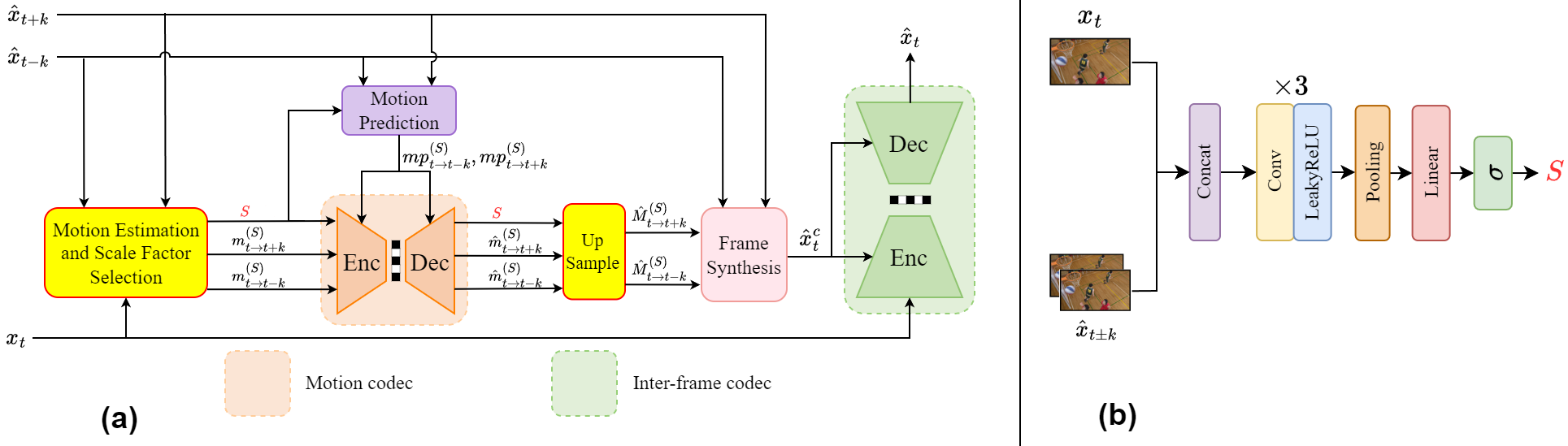}}
    \caption{Illustration of our compression system. (a) The coding framework of MaskCRT B-frame, where $x_t$ is the current coding B-frame and $\hat{x}_{t-k}, \hat{x}_{t+k}$ are the two previously reconstructed reference frames. $S$ represents the downsampling factor, which takes on 1, 2, 4, or 8. (b) The proposed network for Bi-Class and Mu-Class to decide the downsampling factor. It takes $x_t$, $\hat{x}_{t-k}, \hat{x}_{t+k}$ as the inputs to predict the downsampling factor $S$.}
    \label{fig:proposed_method}
\end{figure*}
\vspace{-0.8mm}

\subsection{Preliminary}

Table \ref{tab:ME_complexity} compares the compute requirements for performing full encoding and decoding with the learned codec MaskCRT B-frame~\cite{maskcrt} as well as performing motion estimation for different downsampling factors. The statistics are collected on $1920 \times 1080$ test videos. As shown, the full encoding and decoding of a high-definition video  is expensive. It takes about 2033 kilo multiply–accumulate operations per pixel (kMAC/pixel). About one third of these operations come from motion estimation (ME) at full resolution. Notably, performing ME at full resolution requires more MACs than the combined MACs from performing ME at the other lower resolutions. 


Based on these observations, we propose integrating a deep learning model to predict whether ME should be done at full or low resolution. Another motive of doing this is to address the domain shift issue. Our model offers a fast mode decision mechanism for the pre-trained MaskCRT B-frame codec.

\setlength{\textfloatsep}{5pt}
\begin{table}[t]
\caption{\justifying The breakdown analysis of complexity in terms of the multiply–accumulate operations per pixel.}
\center
\small
\begin{tabular}{ccr}
\hline
\textbf{Module}        & \textbf{kMAC/pixel}   & \textbf{Percentage} \\ \hline
{MaskCRT B-frame}       & 2033    & 100\%               \\
{ME with S = 1} & 645 & 31.7\%             \\
{ME with S = 2} & 161 & 7.9\%              \\
{ME with S = 4} & 40  & 2.0\%              \\
{ME with S = 8} & 11 & 0.5\%              \\ \hline
\end{tabular}
\label{tab:ME_complexity}

\end{table}



\subsection{System Overview}
\label{sssec:system_overview}
We conduct our experiments on the B-frame extension of MaskCRT \cite{maskcrt}, as illustrated in Fig. \ref{fig:proposed_method} (a). To encode an input frame $x_t \in \mathbb{R}^{3 \times W \times H}$ of width $W$ and height $H$, we first select the optimal downsampling factor $S$ from \{1, 2, 4, 8\}. The coding frame and its reference frames are then downsampled accordingly as $x^{(S)}_t, \hat{x}^{(S)}_{t-k}, \hat{x}^{(S)}_{t+k} \in \mathbb{R}^{3 \times \frac{W}{S} \times \frac{H}{S}}$, before performing motion estimation to obtain two low-resolution optical flow maps $m^{(S)}_{t\to{t}-k}, m^{(S)}_{t\to{t}+k} \in \mathbb{R}^{2 \times \frac{W}{S} \times \frac{H}{S}}$. The motion encoder then compresses these estimated flow maps based on the predicted flow maps $mp^{(S)}_{t\to{t}-k}, mp^{(S)}_{t\to{t}+k} \in \mathbb{R}^{2 \times \frac{W}{S} \times \frac{H}{S}}$ generated by the Motion Prediction network as the conditioning signals. The decoded flow maps are subsequently upsampled to the original resolution, as $\hat{M}^{(S)}_{t\to{t}-k}, \hat{M}^{(S)}_{t\to{t}+k} \in \mathbb{R}^{2 \times W \times H}$, and fed into the Frame Synthesis network, along with the two high-resolution reference frames to generate motion compensated frame $mc^{(2)}_t \in \mathbb{R}^{3 \times W \times H}$, which serves as the conditioning signal for the inter-frame codec to encode the coding frame $x_{t}$.

In determining the downsampling factor $S$, OMRA~\cite{OMRA} adopts an exhaustive search strategy via a rate-distortion optimization (RDO) process, which is computationally intensive and time-consuming. To address this, we propose a low-complexity approach termed Fast-OMRA. 






\subsection{Fast-OMRA Implementations}
\label{subsec:variants}
We investigate two implementations of Fast-OMRA: (1) binary-class OMRA (Bi-Class) and (2) multi-class OMRA (Mu-Class). Both methods share a similar classification model (see Fig. \ref{fig:proposed_method} (b)) and take the coding frame $x_{t}$, along with its reference frames $\hat{x}_{t \pm k}$, as input signals to predict the downsampling factor $S$ for each coding frame.

\subsubsection{Binary-class OMRA (Bi-Class)}
We divide the prediction classes into two categories: (1) the static motion class, where the optimal downsampling factor $S$ is set to 1, and (2) the complex motion class, which performs the online selection of $S \in \{ 2, 4, 8 \}$ by performing ME at lower resolutions and evaluating the prediction errors after superresolving the low-resolution optical flow maps for temporal warping in full resolution. This design choice is motivated by the complexity analysis in Table~\ref{tab:ME_complexity}, which shows that performing ME at the highest resolution ($S$=1) alone is more expensive than performing ME at the remaining resolutions (i.e. $S \in \{2, 4, 8 \}$). 

Given that predicting multiple classes including the downsampling factors $S \in \{1, 2, 4, 8 \}$ is a non-trivial task, only two classes are included in this variant. We train multiple classifiers (one for each temporal level in a hierarchical temporal prediction structure). We perform OMRA~\cite{OMRA} to obtain the optimal downsampling factor as the ground truth label for each training sample and apply the Focal Loss~\cite{focal_loss} to balance the class distribution. The training objective is given by:
\begin{equation}
\label{equ:focal_loss}
L_{Bi} = \alpha_{t} \cdot \left ( 1-p_{t} \right )^\gamma \cdot CE(p_t, L_{hard}),
\end{equation}
where $\alpha_t$ is used to balance the uneven distributions of the two prediction classes, $(1-p_t)^\gamma$ is to discount the well-classified samples, $\gamma$ is set empirically to 2, $L_{hard}$ is the ground truth label in the form of a one-hot vector, and CE represents the cross-entropy loss. $p_t$ is set to the classifier's output $p$ if the ground truth label is 1 and $1-p$ otherwise.

\subsubsection{Multi-class OMRA (Mu-Class)}
With this variant of Fast-OMRA, we predict four classes corresponding to the four different downsampling factors $S$ to further reduce complexity. The predicted downsampling factor is directly used to encode the video frames, eliminating the need for an exhaustive search of the optimal factor. However, the input signals to the classifier are weak state signals, which contain only intensity information without any temporal information ($x_t$ and $\hat{x}_{t \pm k}$), making it challenging for the model to make accurate predictions. 
Inspired by knowledge distillation techniques, we propose a novel soft label training strategy that softens the ground truth label to more closely reflect the relationship among the rate-distortion (RD) costs resulting from these downsampling factors. Specifically, we soften the ground truth label as follows:
\begin{equation}
\label{equ:focal_loss}
L_{soft} = \frac{e^{\lambda a_{i}}}{\sum e^{\lambda a_{i}}},
\end{equation}
\begin{equation}
\label{equ:focal_loss}
 a_{i}= \frac{RD_{max}-RD_{i}}{RD_{max}},
\end{equation}
where $RD_i$ is the RD cost from the use of a specific downsampling factor and $RD_{max}$ is the largest RD cost among the four downsampling factors. It is seen that a smaller $\lambda$ leads to a softer ground truth and vice versa. With this design, if the four downsampling factors give rise to similar RD costs, the model should output similar probabilities for them accordingly. In the present case $\lambda$ is set to 10.

Recognizing that ambiguous training samples, where the four different downsampling factors yield similar RD costs, can confuse the model during training, we aim to discount these training samples and have the model focus more on predicting well on distinct cases. To achieve this, an entropy measure of the softened ground truth label $L_{soft}$ is introduced to distinguish between ambiguous and distinct samples. When a sample is ambiguous, the softened label approaches 0.25 for each class, resulting in an entropy value close to 2, and such training samples are discounted. We construct the training objective as follows:
\begin{equation}
\label{equ:focal_loss}
L_{Mu} = \left ( 2-Entropy(L_{soft}) \right ) \cdot CE(p_t, L_{soft}),
\end{equation}
where $L_{soft}$ is the softened ground truth label. 

\begin{figure*}[t]
    \begin{center}
    \vspace{-1.0em}
    \begin{subfigure}{0.32\linewidth}
        \centering
        \includegraphics[width=\linewidth]{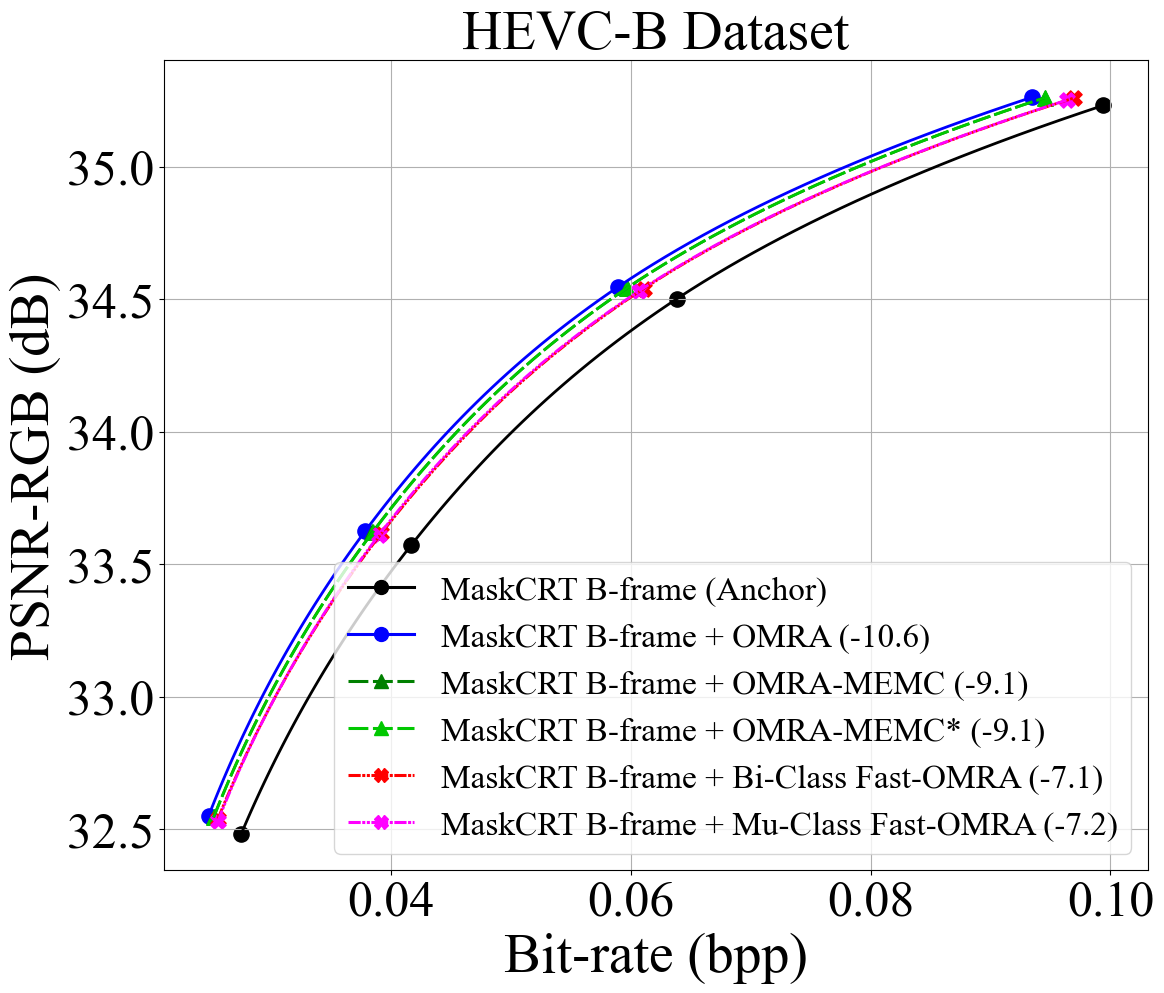}
        \label{fig:hevcb}
    \end{subfigure}
    \begin{subfigure}{0.327\linewidth}
        \centering
        \includegraphics[width=\linewidth]{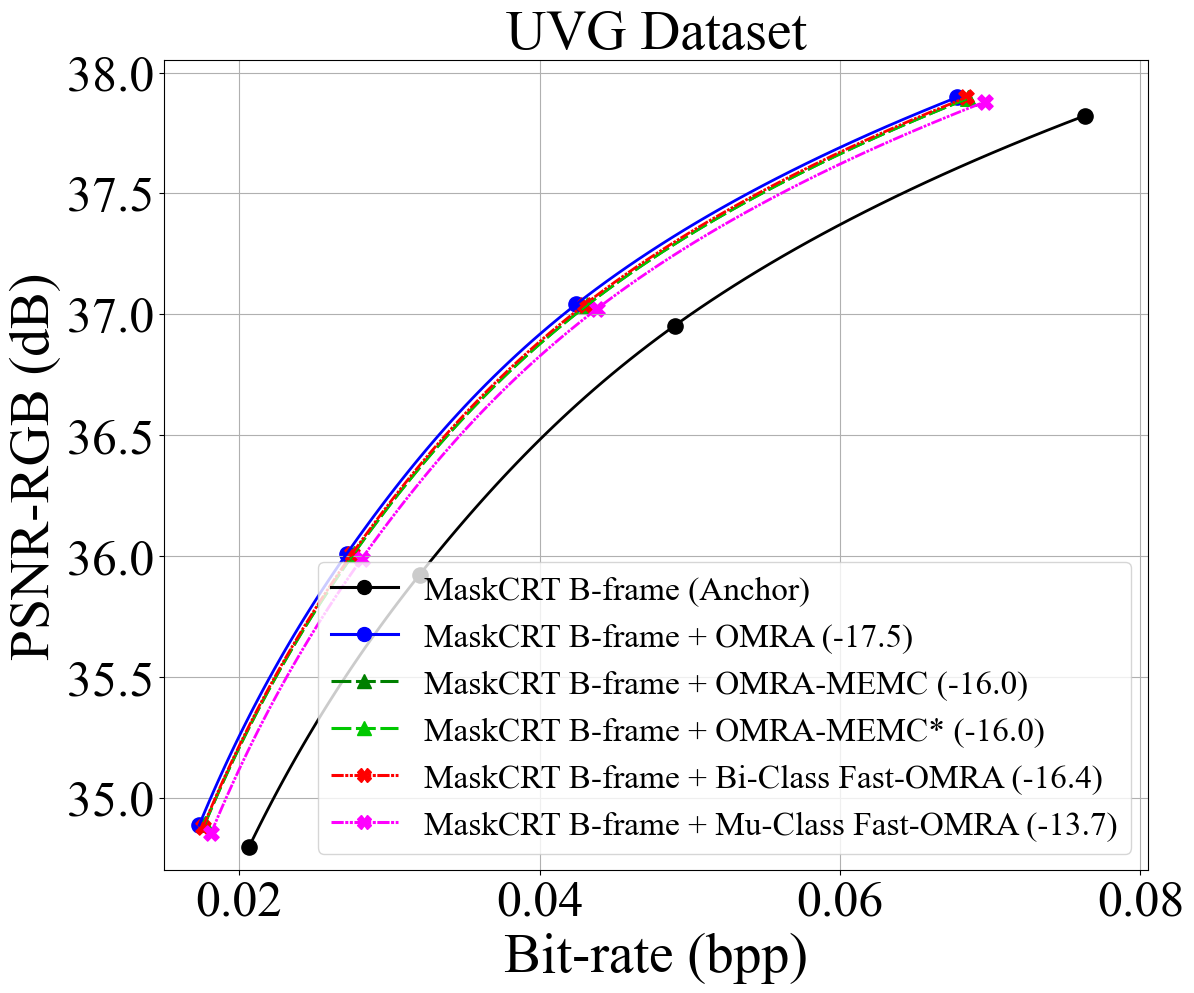}
        \label{fig:uvg}
    \end{subfigure}
    \begin{subfigure}{0.32\linewidth}
        \centering
        \includegraphics[width=\linewidth]{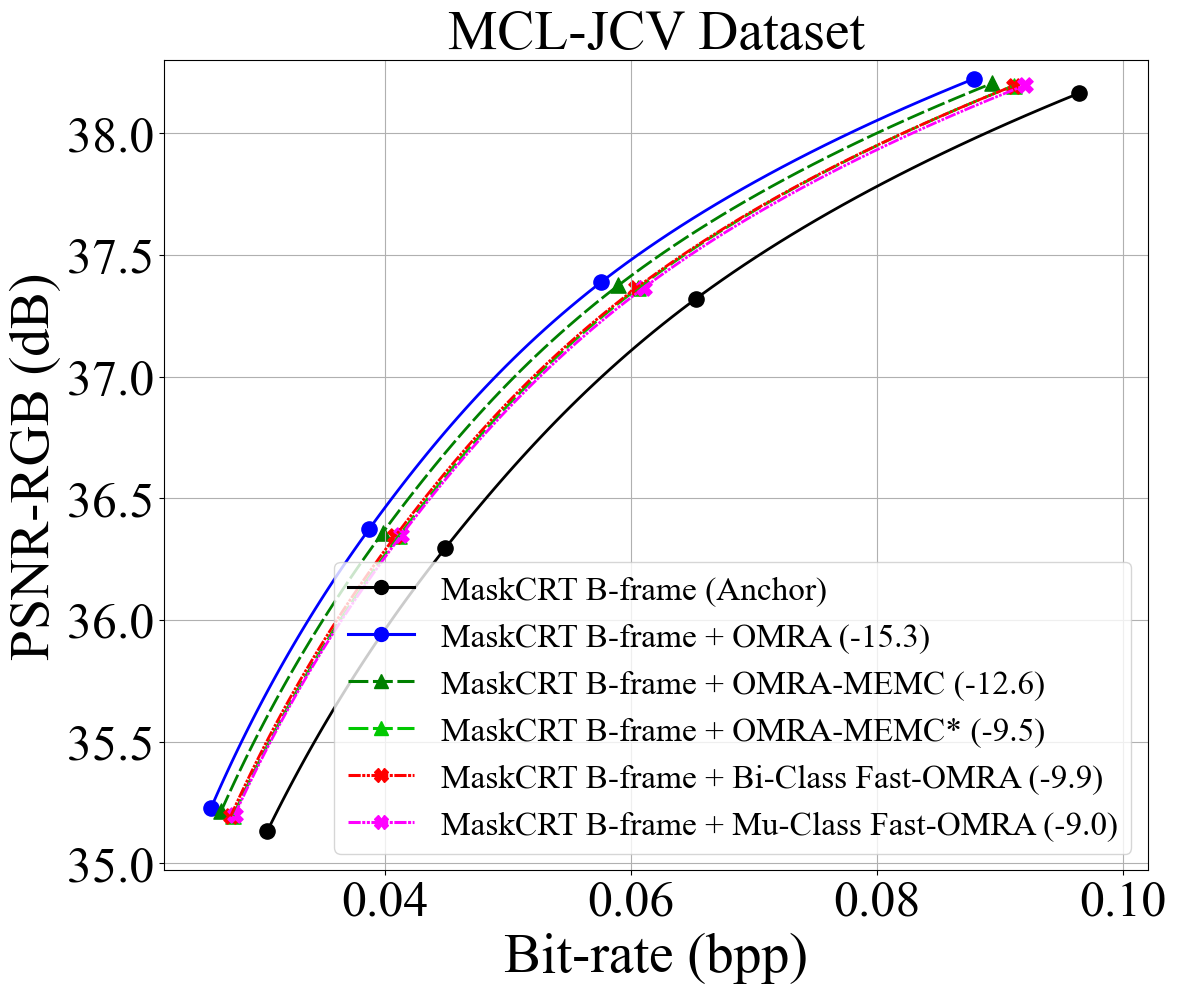}
        \label{fig:mcl}
    \end{subfigure}
    \vspace{-1.0em}
    \caption{The rate-distortion performance comparison. The anchor is MaskCRT B-frame.}
    \label{fig:RD}
    \end{center}
    \vspace{-1.0em}
\end{figure*}
\begin{figure}[t]
    \centering
    \vspace{-0.8em}
    \centerline{\includegraphics[width=0.48\textwidth]{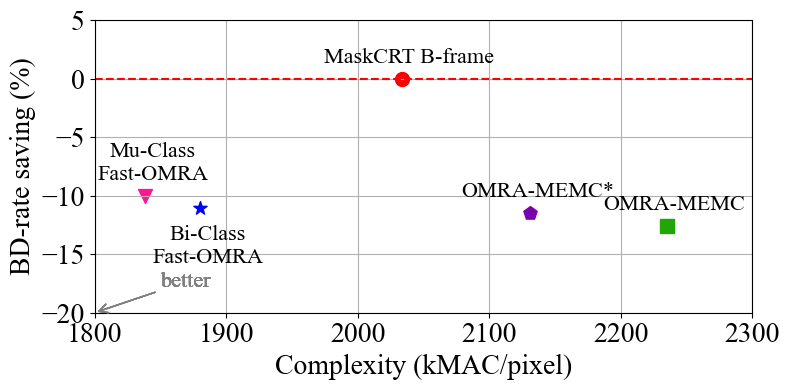}}
    
    \caption{The rate-distortion-complexity trade-offs.}
    \label{fig:rd_complexity}
\end{figure}
\section{Experimental Results}
\label{sec:results}
In this section, we evaluate the performance of the presented methods in terms of both coding efficiency and computational complexity. OMRA-MEMC is defined as a low-complexity variant of OMRA, where instead of performing the full RDO process, it only conducts ME and backward warping (i.e. motion compensation), and then compares the prediction errors to select the optimal downsampling factor. OMRA-MEMC* further reduces complexity by directly using $S = 1$ at the highest temporal level without testing other downsampling factors.
\subsection{Settings}
\label{ssec:setting}
For training our classifiers, we use the BVI-DVC \cite{BVI} and TVD \cite{TVD} datasets, and adopt the Random Path sampling strategy \cite{longvideocoding} with the GOP size 32. Training sequences are randomly cropped into 416 × 256 patches. The proposed method is evaluated on the HEVC Class B \cite{hevcctc}, UVG \cite{uvg} and MCL-JCV \cite{mcl} datasets. Each test video sequence is encoded with 97 frames, with the intra-period and GOP size set to 32. We adopt a hieararchical B-frame coding structure $I_{0}B_{5}B_{4}B_{5}...B_{2}......B_{1}.........I_{0}$ to organize the video frames in a GOP into 5 temporal levels (with the temporal level indicated in the subscript). We train 1 single classifier or 4 separate classifiers for the 4 lowest temporal levels (i.e. the temporal level from 1 to 4), while at the highest temporal level (i.e. the temporal level 5), we choose $S = 1$. We report the BD-rate saving~\cite{bdrate} of the competing methods with the anchor being MaskCRT B-frame.


\subsection{Experiments}
\label{ssec:ablation}
\subsubsection{Cross Entropy Loss vs. Focal Loss}
\label{sssec:ablation_celoss}
To justify the use of Focal Loss, we train our Bi-Class model using the popular Cross Entropy (CE) Loss. From Table~\ref{tab:ablation_binary}, we see that Focal Loss performs much better than CE Loss. The reason is that CE Loss treats all samples equally, which may not be ideal in cases where some frames are more challenging to decide which downsampling factor to use. In comparison, Focal Loss focuses more on hard-to-classify samples by using $ \left ( 1-p_{t} \right )^\gamma$ as the focusing weight.
\begin{table}[t]
\caption{The loss type and number of models for Bi-Class}
\centering
\scriptsize
\setlength{\tabcolsep}{5pt}

\begin{tabular}{lccccc}
\hline
Bi-Class            & HEVC-B & UVG   & MCL-JCV & \begin{tabular}[c]{@{}c@{}}Average\\ BD-rate\end{tabular} & \begin{tabular}[c]{@{}c@{}}Average\\ kMAC/pixel\end{tabular} \\ \hline
1 model, CE Loss    & -2.6   & -6.4  & -3.8    & -4.3                                                      & 1979                                                  \\
1 model, Focal Loss & -3.6   & -16.4 & -9.2    & -9.7                                                      & 1883                                                  \\
4 models, Focal Loss & -7.1   & -16.4 & -9.9    & -11.1                                                     & 1880                                                  \\ \hline
\end{tabular}
\label{tab:ablation_binary}

\end{table}

\subsubsection{Soft Ground Truth Labels}
\label{sssec:softlabel}
Table~\ref{tab:ablation_softlabel} compares the effectiveness of our soft labels against hard labels, and examines the impact of using $\left ( 2-Entropy(L_{soft}) \right )$ as the weighting scheme in the training objective. The results indicate that softened labels, which incorporate information about the RD costs of the 4 downsampling factors, help improve prediction accuracy.  Additionally, applying the weighting $\left ( 2-Entropy(L_{soft}) \right )$ helps the model focus more on distinct cases, where wrong predictions could lead to more considerable coding loss. 
\begin{table}[t]
\caption{The loss type and number of models for Mu-Class}
\centering
\scriptsize
\setlength{\tabcolsep}{3.6pt}
\begin{tabular}{lccccc}
\hline
Mu-Class                        & HEVC-B & UVG  & MCL-JCV & \begin{tabular}[c]{@{}c@{}}Average\\ BD-rate\end{tabular} & \begin{tabular}[c]{@{}c@{}}Average\\ kMAC/pixel\end{tabular} \\ \hline
1 Model, Hard             & -6.6   & -11.3 & -8.4    & -8.8                                                      & 1848                                                   \\
1 Model, Soft w/o Entropy & -7.6   & -12.0 & -8.7    & -9.4                                                      & 1858                                                   \\
1 Model, Soft w/ Entropy  & -7.2   & -13.7 & -9.0    & -10.0                                                     & 1838                                                  \\
4 Models, Soft w/ Entropy  & -7.5   & -11.9 & -9.0    & -9.5                                                      & 1851                                                   \\ \hline
\end{tabular}
\label{tab:ablation_softlabel}
\end{table}

\subsubsection{Single Model vs. Multiple Models}
\label{sssec:multimodel}
From Table~\ref{tab:ablation_binary}, having 4 separate Bi-Class classifiers to classify on video frames at each of the lowest 4 temporal levels improves coding performance. However, in Table~\ref{tab:ablation_softlabel}, incorporating more Mu-Class classifiers decreases the coding performance slightly on UVG. Admittedly, this deserves further investigation.

\subsection{Comparison of Rate-Distortion-Complexity Performance}
\label{ssec:Compresison_performance}
\begin{table}[t]
\centering
\caption{Comparison of the BD-rate saving. The anchor is MaskCRT B-frame.}
\scriptsize
\begin{tabular}{ccccc}
\hline
Dataset             & HEVC-B  & UVG      & MCL-JCV & Average          \\ \hline
OMRA                & -10.6 & -17.5  & -15.3 & -14.5 \\
OMRA-MEMC           & -9.1  & -16.0  & -12.6 & -12.6          \\
OMRA-MEMC*           & -9.1  & -16.0  & -9.5  & -11.5          \\
Bi-Class Fast-OMRA & -7.1  & -16.4 & -9.9 & -11.1         \\
Mu-Class Fast-OMRA & -7.2 & -13.7 & -9.0 & -10.0 \\
\hline
\end{tabular}
\label{tab:performce}

\end{table}
\begin{table}[t]
\centering
\caption{Comparison of kMAC/pixel. Our Fast-OMRA has content-dependent kMAC/pixel because of its non-exhaustive search nature in determining the best downsampling factor.}
\scriptsize
\begin{tabular}{ccccc}
\hline
Dataset             & HEVC-B  & UVG     & MCL-JCV & Average \\ \hline
OMRA                & \multicolumn{4}{c}{5863}           \\
OMRA-MEMC           & \multicolumn{4}{c}{2235}            \\
OMRA-MEMC*           & \multicolumn{4}{c}{2131}             \\
Bi-Class Fast-OMRA & 1893 & 1863 & 1885 & 1880 \\  
Mu-Class Fast-OMRA & 1872 & 1820 & 1822 & 1838 \\
\hline
\end{tabular}
\label{tab:complexity}
\vspace{-0.5em}
\end{table}
The rate-distortion performance of the competing methods is depicted in Table~\ref{tab:performce} and Fig. \ref{fig:RD}, while Table \ref{tab:complexity} presents the kMAC/pixel needed for each method and Fig. \ref{fig:rd_complexity} characterizes their rate-distortion-complexity trade-offs. Hereafter, Bi-Class refers to its 4-model variant plus Focal Loss, whereas Mu-Class is the 1-model variant trained with soft labels plus entropy weighting. 

From the results, using motion-compensated prediction errors to determine the optimal downsampling factor, as with OMRA-MEMC, reduces the complexity from 5863 kMAC/pixel with OMRA to 2235 kMAC/pixel, while still maintaining high compression performance (12.6\% BD-rate saving, as opposed to 14.5\% with OMRA). OMRA-MEMC*, which chooses $S=1$ for video frames at the highest temporal level, also demonstrates high compression performance with 11.5\% BD-rate saving and reduces the computational complexity to 2131 kMAC/pixel.

Our proposed Bi-Class and Mu-Class Fast-OMRA shows much improved coding performance than MarkCRT B-frame (the anchor) across all three datasets, achieving 11.1\% and 10.0\% average BD-rate saving, respectively. It indicates that our approach effectively mitigates the negative impact of domain shift. This is largely due to the model's ability to accurately predict the use cases of low-resolution ME, particularly for frames with fast motion and low temporal correlation. Notably, Fast-OMRA variants achieve even lower complexity than MarkCRT B-frame (1880 and 1838 kMAC/pixel, respectively). This is because MarkCRT B-frame consistently uses full-resolution ME for encoding, whereas our Fast-OMRA encodes a significant number of frames using low-resolution ME. Our Bi-Class and Mu-Class models share a similar model size of 0.37M parameters. This extra network cost is marginal as compared to MaskCRT B-frame, which has a model size of 34.9M.

\section{Conclusion}
In this paper, we present a low-complexity approach to addressing the domain shift issue for learned B-frame codecs. Based on the finding that downsampling input video frames can help the model predict the motion more accurately, we train lightweight classifiers to decide the best scale factor as an alternative to performing an exhaustive full RDO search. We experiment with two variants, Bi-Class and Mu-Class. The former leverages our complexity analysis to classify between ME at high and low resolutions, while the latter skips entirely the search of the scale factors. Soft labels are used in training Mu-Class to account for the weak state inputs. The experiment results show that both variants strike a good trade-off among rate, distortion, and complexity.


\bibliographystyle{IEEEtran}
\bibliography{IEEEabrv,paper.bib}
\end{document}